\documentclass[conference]{IEEEtran}
\IEEEoverridecommandlockouts
\usepackage{color}

\usepackage{cite}
\usepackage{amsmath,amssymb,amsfonts}
\usepackage{algorithmic}
\usepackage{graphicx}
\usepackage{textcomp}
\usepackage{xcolor}
\usepackage[bookmarks=false]{hyperref}
\usepackage{booktabs}
\usepackage{multirow}
\usepackage{romannum}
\usepackage[caption=false, labelfont=sf, textfont=sf]{subfig}
\usepackage[normalem]{ulem}
\useunder{\uline}{\ul}{}
\def\BibTeX{{\rm B\kern-.05em{\sc i\kern-.025em b}\kern-.08em
    T\kern-.1667em\lower.7ex\hbox{E}\kern-.125emX}}
\makeatletter
\def\UrlAlphabet{%
      \do\a\do\b\do\c\do\d\do\e\do\f\do\g\do\h\do\i\do\j%
      \do\k\do\l\do\m\do\n\do\o\do\p\do\q\do\r\do\s\do\t%
      \do\u\do\v\do\w\do\x\do\y\do\z\do\A\do\B\do\C\do\D%
      \do\E\do\F\do\G\do\H\do\I\do\J\do\K\do\L\do\M\do\N%
      \do\O\do\P\do\Q\do\R\do\S\do\T\do\U\do\V\do\W\do\X%
      \do\Y\do\Z}
\def\UrlDigits{\do\1\do\2\do\3\do\4\do\5\do\6\do\7\do\8\do\9\do\0}
\g@addto@macro{\UrlBreaks}{\UrlOrds}
\g@addto@macro{\UrlBreaks}{\UrlAlphabet}
\g@addto@macro{\UrlBreaks}{\UrlDigits}
\makeatother

\graphicspath{{media/}}
\begin{document}

\title{Training-free Monocular 3D Event Detection System for Traffic Surveillance
\thanks{This project is funded in part by Carnegie Mellon University’s Mobility21 National University Transportation Center, which is sponsored by the US Department of Transportation. This research is supported in part through the financial assistance award 60NANB17D156 from U.S. Department of Commerce, National Institute of Standards and Technology and by the Intelligence Advanced Research Projects Activity (IARPA) via Department of Interior/Interior Business Center (DOI/IBC) contract number D17PC00340.
}
}

\author{\IEEEauthorblockN{Lijun Yu}
\IEEEauthorblockA{\textit{Carnegie Mellon University}\\
Pittsburgh, PA, US \\
lijun@cmu.edu}
\and
\IEEEauthorblockN{Peng Chen}
\IEEEauthorblockA{\textit{Peking University}\\
Beijing, China \\
chen.peng@pku.edu.cn}
\and
\IEEEauthorblockN{Wenhe Liu}
\IEEEauthorblockA{\textit{Carnegie Mellon University}\\
Pittsburgh, PA, US \\
wenhel@andrew.cmu.edu}
\and
\IEEEauthorblockN{Guoliang Kang}
\IEEEauthorblockA{\textit{Carnegie Mellon University}\\
Pittsburgh, PA, US \\
gkang@andrew.cmu.edu}
\and
\IEEEauthorblockN{Alexander G. Hauptmann}
\IEEEauthorblockA{\textit{Carnegie Mellon University}\\
Pittsburgh, PA, US \\
alex@cs.cmu.edu}
}

\maketitle

\begin{abstract}
    We focus on the problem of detecting traffic events in a surveillance scenario, including the detection of both vehicle actions and traffic collisions.
    Existing event detection systems are mostly learning-based and have achieved convincing performance when a large amount of training data is available.
    However, in real-world scenarios, collecting sufficient labeled training data is expensive and sometimes impossible (e.g. for traffic collision detection).
    Moreover, the conventional 2D representation of surveillance views is easily affected by occlusions and different camera views in nature.
    To deal with the aforementioned problems, in this paper, we propose a training-free monocular 3D event detection system for traffic surveillance.
    Our system firstly projects the vehicles into the 3D Euclidean space and estimates their kinematic states.
    Then we develop multiple simple yet effective ways to identify the events based on the kinematic patterns, which need no further training.
    Consequently, our system is robust to the occlusions and the viewpoint changes.
    Exclusive experiments report the superior result of our method on large-scale real-world surveillance datasets, 
    which validates the effectiveness of our proposed system.
    The demonstration videos of our system are available online\footnote{\url{https://drive.google.com/drive/folders/118tdbpWhfJC-7tT9whzEsHPWjx6hdEi7?usp=sharing}}.
\end{abstract}


\section{Introduction}

Nowadays, traffic surveillance cameras are broadly installed for safety purposes, recording massive data continuously. 
Due to the large volume and high updating velocity of data, assistance of artificial intelligence is demanded for data distillation and analysis of the surveillance data.
Researchers have designed well-performing event detection systems with supervised learning methods.
However, when meeting real-world scenarios, these systems demonstrate obvious limitations.
First, most of the previous systems rely on massive training data.
In order to learn to detect and distinguish different types of the events, high quality and large number annotations for training data is indispensable.
For some events, such as for rare traffic collision events, e.g., car crashes, it could be very challenging to collect a large enough training dataset.
Moreover, data annotation requires a huge investment of time and labor resource, which is usually limited in real-world.
Second, even given enough annotations, directly processing the two-dimensional format of the surveillance videos is still difficult. The capability of the systems will be limited due to the common problems of 2D video analysis, e.g.,  occlusions and different viewing angle of the camera.
More specifically, for the traffic events, as a result of missing depth information, the speed and location of vehicles cannot be estimated precisely in 2D videos. Unfortunately, it is more expensive to setup the 3D/RGB-D cameras than 2D cameras in real-world.

\begin{figure}[!t]
    \centering
    \includegraphics[width=0.6\linewidth]{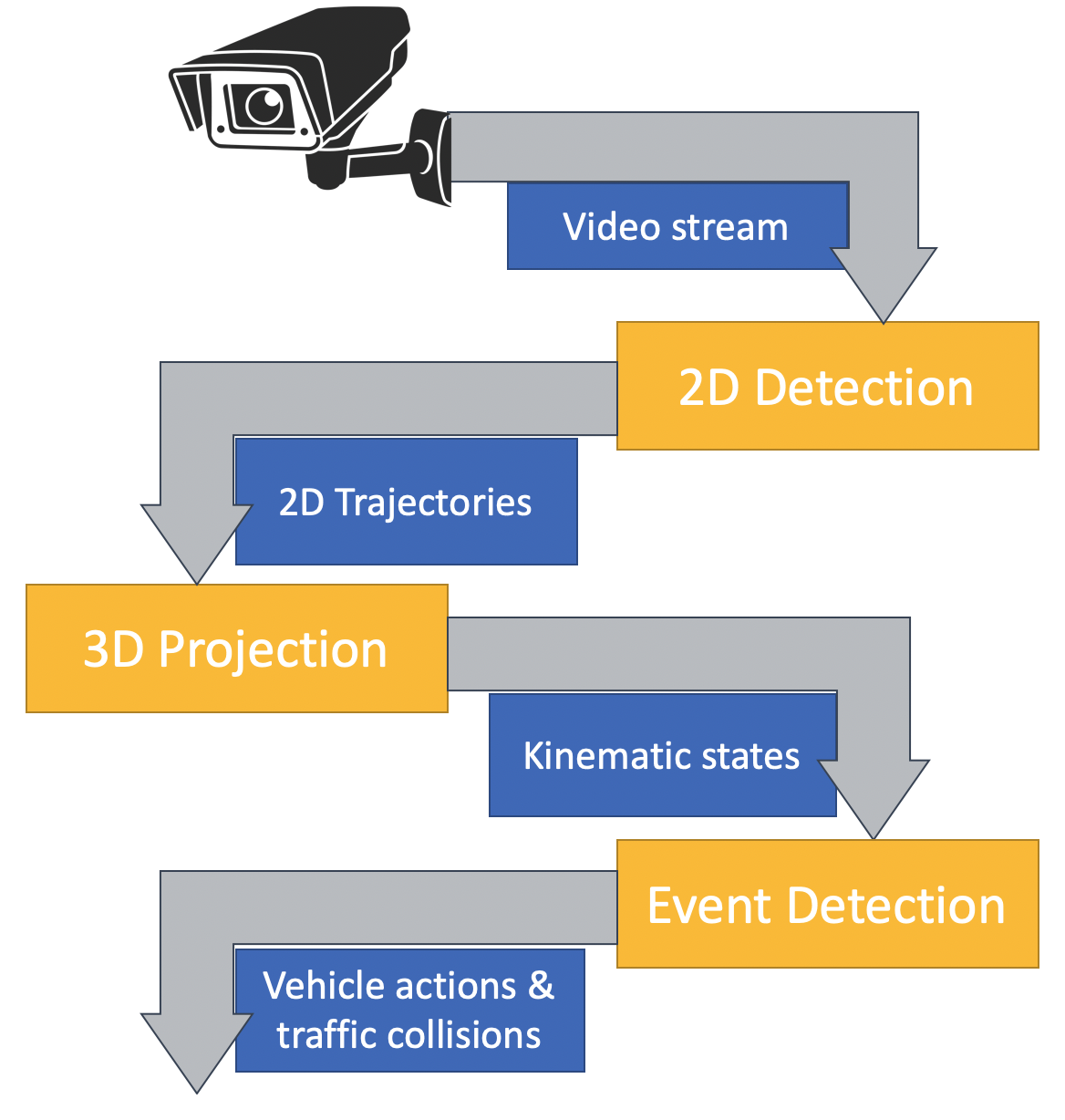}
    \caption{System Overview}
    \label{fig/intro}
\end{figure}

To deal with the problems mentioned above, we developed a monocular 3D event detection system for traffic surveillance, which is shown in Fig. \ref{fig/system}.
In this system, we firstly generate 2D trajectories via object detection and tracking on video stream from single 2D cameras.
Then, with camera calibration parameters, the 2D trajectories are projected into the 3D Euclidean space. 
After that, we are able to measure the kinematic state of each vehicle directly in a top-down view of the road. Therefore, it naturally overcomes the occlusion problems and avoids the difference of viewing angles.
We propose an event detection model with no need for labeled training data, which is based on the kinematic patterns of vehicles.
This model can detect vehicle actions such as turning and stopping, and also traffic collisions.
Furthermore, the system is optimized to support real-time stream processing and to deal with large-scale data in real-world  surveillance scenarios.

We state the major contributions of our work as follows:
\begin{enumerate}
    \item We propose a traffic event detection system with no need of training data.
    To the best of our knowledge, this is the first attempt of a training free system for large-scale traffic event detection.
    \item We propose a monocular 3D method for event detection for the traffic surveillance. It is more robust to the occlusions and camera viewing angle changes, compared to the conventional 2D methods.
    \item Our system supports real-time stream processing for large-scale traffic data. Moreover, it outperforms the competitive training-based baseline\cite{chen2019minding} on the challenging real-world surveillance dataset.
\end{enumerate}

The rest of this work is organized as follows: 
In section \Romannum{2} we first revisit the former systems related to monocular 3D surveillance and then propose our system.
In Section \Romannum{3} we revisit the existing event detection approaches and then introduce our training-free model.
In Section \Romannum{4} we cover the details of the experiments.

\section{Monocular 3D Surveillance}

Most of the event detection methods process in the two-dimensional format of surveillance view, which lacks the depth information.
A latest traffic danger recognition model \cite{yu2018traffic} utilized the 3D projection with calibrated cameras, but only supported vehicles moving in straight roads.
In this paper, we develop a monocular 3D approach that supports vehicles moving in arbitrary directions.
This enables us to detect vehicle turning actions and handle complicated intersections of roads.

\begin{figure}[!t]
    \centering
    \includegraphics[width=0.65\linewidth]{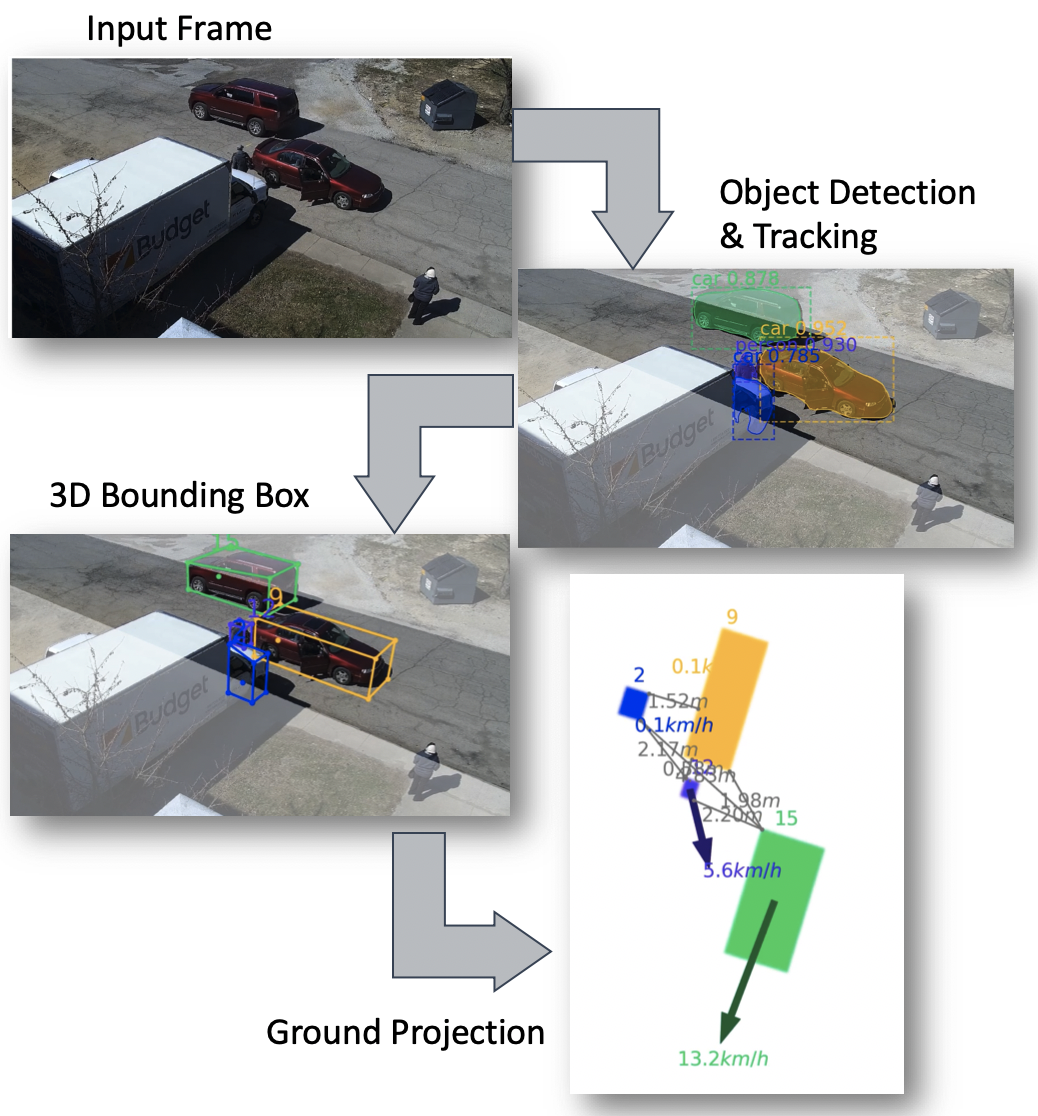}
    \caption[Monocular 3D Surveillance Overview]{Monocular 3D Surveillance Overview \\Through 2D detection and 3D projection, the kinematic state of each vehicle is obtained in a top-down view.}
    \label{fig/m3dis}
\end{figure}

From a single view of a surveillance camera, our system can get the kinematic state of each vehicle in the 3D world space.
A brief sample of the monocular 3D surveillance pipeline is shown in Fig. \ref{fig/m3dis}.

\subsection{2D Detection}

For input video, detection and tracking of vehicles are first done in the 2D image space.

\subsubsection{Image Object Detection}

We utilize the state-of-the-art model Mask R-CNN\cite{he2017mask} for frame-level object detection. 
Apart from the original outputs of object types, detection scores, 2D bounding boxes, and object masks, we further extract the Region-of-Interest features from the feature pyramid network\cite{lin2017feature} for tracking.

We adopted the object detection implementation from \cite{chen2019minding}.
Although fine-tuning on specific datasets may improve the detection performance, it requires annotations of objects.
Since many video datasets do not contain such annotations, we used the weights trained on COCO\cite{lin2014microsoft} to provide a generalized ability of object detection without additional training.
In this task, only objects classified as car, bus, or truck are selected for tracking.

\subsubsection{Online Object Tracking}

With the deep feature from our detection model, we employ the state-of-the-art online tracking algorithm Deep SORT\cite{wojke2017simple}.
It tracks the detected vehicles and handles brief lost of detection.
A Kalman filter with a constant velocity model is used to estimate the location and speed of objects in the image space from the noisy 2D bounding boxes.

Although offline tracking algorithms may achieve better performance, we did not pursue them here as they are not suitable for the online realtime goal of a surveillance system.

\subsection{3D Projection}

With the transformation parameters acquired from camera calibration, we can project the detected and tracked vehicles from the 2D image onto the ground in the 3D space.

\subsubsection{Camera Calibration}

We adopt a road surveillance camera model as in \cite{sochor2017traffic}.
According to \cite{hartley2003multiple}, the relationship between the homogeneous coordinates of a point $\mathbf{X} = \begin{pmatrix} X \\ Y \\ Z \end{pmatrix}$ in the Euclidean 3D world space and its projection $\mathbf{x} = \begin{pmatrix} x \\ y \end{pmatrix}$ in the 2D image space follows 
\begin{equation}
    \begin{pmatrix} x \\ y \\ 1 \end{pmatrix} = K [R | t] \begin{pmatrix} X \\ Y \\ Z \\ 1 \end{pmatrix} = P \begin{pmatrix} X \\ Y \\ Z \\ 1 \end{pmatrix}
\end{equation}
where $P = K [R | t]$ is the projection matrix. $K$ contains the internal camera parameters. $R$ and $t$ provides the rotation and translation vector between world and camera coordinates.

Some datasets provide the $K, R, t$ parameters of each camera calibrated at installation, which can be used directly.
Some other datasets provide two vanishing points as calibration.
If nothing is provided, we can manually label two sets of parallel lines perpendicular to each other for each camera view.
Then we can derive two vanishing points with a method\cite{lee2011robust} based on the least square error.
With 2 vanishing points $u, v$, we can get $K, R, t$ according to \cite{dubska2014fully}.

With the projection matrix $P$, we can reproject any image point onto a known plane in the 3D space, such as the ground plane.

\subsubsection{3D Bounding Box}

For a detected vehicle, we estimate its contour from its mask using the algorithm in \cite{suzuki1985topological}.
We get the speed vector of the bottom middle point of its 2D bounding box from the internal state of the Kalman filter.
Projecting this vector onto the ground, we know the vehicle speed in world coordinates.

We acquire three vanishing points $u, v, w$ on the image according to its speed.
$u$ is in its moving direction, $v$ is on the ground and perpendicular to $u$, and $w$ is perpendicular to the ground. 
Then we can find the tangent lines of the contour passing these vanishing points and build a 3D bounding box as described in \cite{yu2018traffic}.

The bottom points of the 3D bounding box can then be reprojected to the ground as the location of the vehicle.
A known limitation is that the direction estimation from 2D speed is unavailable for stopped vehicles.
So their 3D bounding boxes could be in an inaccurate orientation.

\section{Training-free Event Detection}

Existing event detection approaches for traffic surveillance are mainly based on supervised learning, such as multiple instance learning \cite{sultani2018real}, motion pattern based learning \cite{xu2018dual}, and Recurrent Neural Network\cite{chen2019minding}.
All of them require sufficient amount of high quality training data to achieve a reasonable performance.
However, such data are often unavailable in real-world applications.
Therefore, we are pursuing a training free system in this paper.

With the estimated states from the monocular 3D surveillance system, we can further detect driving events and vehicle crashes based on kinematic patterns. 
All these detectors require no annotated training data.

\subsection{Vehicle Action Detection}

In this work, we define and annotate the MEVA dataset with the following vehicle actions. 
The definition of the events are listed below:

\begin{itemize}[\IEEEsetlabelwidth{1000}]
    \item Vehicle turning left: A vehicle turning left from running straight.
    \item Vehicle turning right: A vehicle turning right from  running straight.
    \item Vehicle U-turn: A vehicle making a U-turn. A U-turn is defined as a 180 degree turn and it gives the appearance of a “U”. The vehicle may not stop in the middle of the event.
    \item Vehicle starting: A vehicle starting from stop. In the event, the vehicle may accelerate from stop.
    \item Vehicle stopping: A vehicle stopping from running. The events starts when the vehicle begins noticeably slowing down.
\end{itemize}

To analyze the vehicle actions, we estimate the states of the detected vehicles. 
Specifically, we calculate the ground speed vector and acceleration of the vehicles via their trajectories in 3D Euclidean Space.

\subsubsection{State Estimation}
We define the \textit{ground speed vector} of a vehicle as 
\begin{equation}
    \mathbf{v} = \begin{pmatrix} v_x \\ v_y \end{pmatrix}
\end{equation}
It is transformed into a polar coordinate system as
\begin{equation}
    \hat{\mathbf{v}} = \begin{pmatrix} v_r \\ v_\theta \end{pmatrix} = \begin{pmatrix} \sqrt{v_x^2 + v_y^2} \\ \text{atan2}(v_y, v_x) \end{pmatrix}
\end{equation}

Let $a_r$ and $a_\theta$ denote the derivatives of $v_r$ and $v_\theta$, where $a_r$ represents the acceleration value and $a_\theta$ the angular acceleration.
For each time point $t=t_0$, we observe $v_r$ and $v_\theta$ in a sliding window of size $2w + 1$, i.e. $t_0 - w, t_0 - w + 1, \hdots, t_0 + w$.
We estimate $a_{r, t_0}$ and $a_{\theta, t_0}$ with simple linear regression models defined as 
\begin{align}
    v_{r, t} &= \beta_r + a_{r, t_0} t \\
    v_{\theta, t} &= \beta_\theta + a_{\theta, t_0} t
\end{align}
With ordinary least squares (OLS) method, we can get the estimated $a_{r, t_0}$ and $a_{\theta, t_0}$.

\subsubsection{Detection Model}

\begin{figure}[!h]
    \centering
    \includegraphics[width=\linewidth]{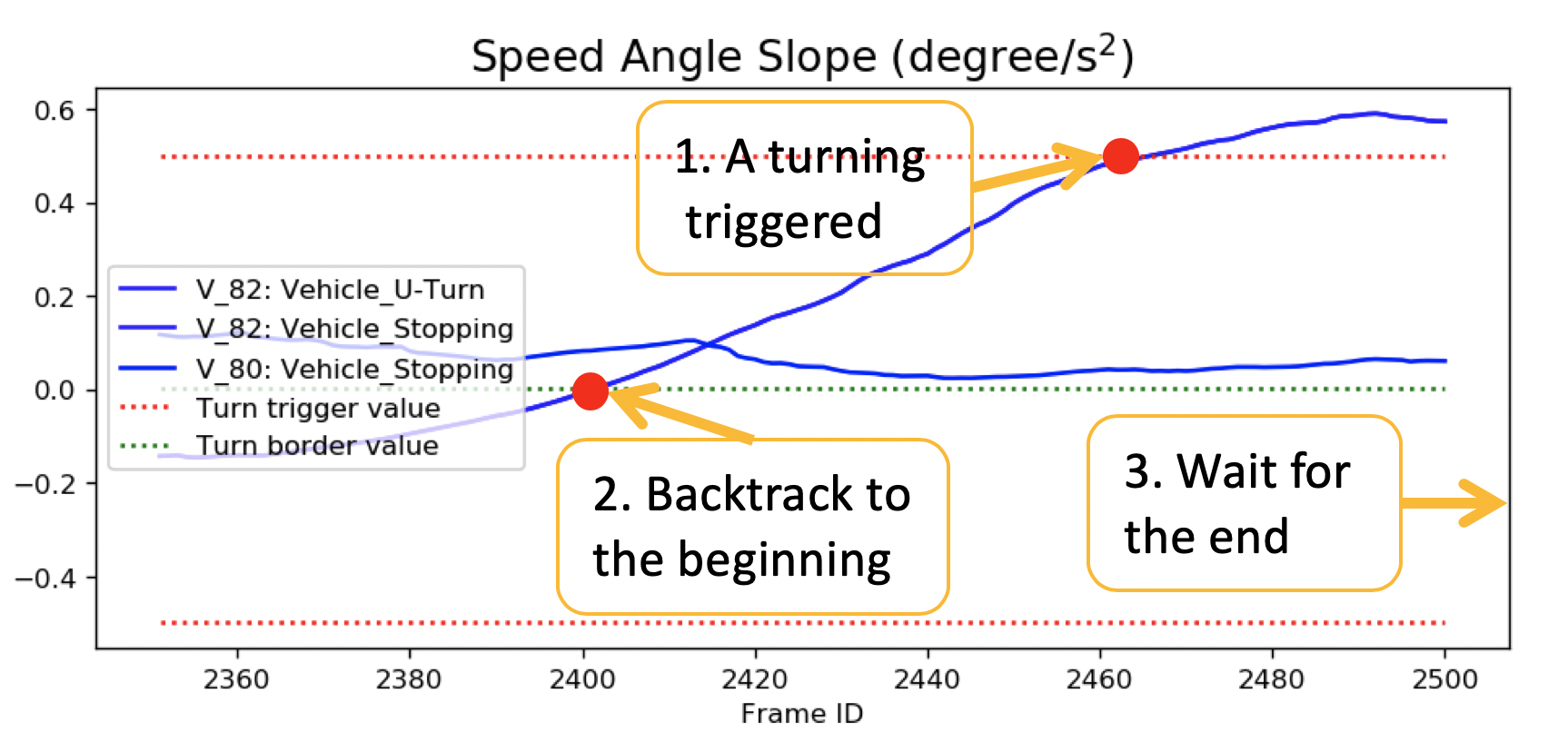}
    \caption[A Sample of Trigger-driven detection for Turning Events]{A Sample of Trigger-driven detection for Turning Events. \\The $y$-axis is $\alpha_\theta$, i.e., the slope of the regression model for speed angle.}
    \label{fig/event}
\end{figure}

The five events are further divided into two groups: the turning events and the linear events.
One trigger-driven event detection model is designed for each group.
A sample of detected turning events is shown in Fig. \ref{fig/event}.

Each model has a frame-level condition of trigger and border, where the latter is looser than the former, and an event-level valid condition.
Once an event is triggered, the model backtracks past frames until the border condition becomes false to decide the beginning of the event.
Then it waits for the future until the border becomes false again to decide the end of the event.
Then it filters the event with the valid condition and applies further classification.

\subsubsection{Turning Events}

The trigger and border conditions of turning events are 
\begin{align}
    |a_\theta| &\ge a_{\theta, trigger} \text{ and } v_r \ge v_{turn\_min} \\
    |a_\theta| &\ge a_{\theta, border} \text{ and } v_r \ge v_{turn\_min}
\end{align}
where $a_{\theta, trigger} > a_{\theta, border}$.
For a turning event started at $t_s$ and ended at $t_e$, the turning angle can be calculated as 
\begin{equation}
    \theta = v_{\theta, t_e} - v_{\theta, t_s}
\end{equation}
where $\theta \in (-180, 180]$.
The valid condition of turning events is
\begin{equation}
    t_e - t_s \ge t_{turn\_min} \text{ and } |\theta| > \theta_{min}
\end{equation}
Then these events are further classified as 
\begin{itemize}[\IEEEsetlabelwidth{1000}]
    \item Vehicle turning left, if $\theta_{min} < \theta < \theta_{max}$
    \item Vehicle turning right, if $-\theta_{max} < \theta < -\theta_{min}$
    \item Vehicle U-turn, if $\theta > \theta_{max}$ or $\theta < -\theta_{max}$
\end{itemize}

\subsubsection{Linear Events}

The trigger and border conditions of linear events are 
\begin{align}
    |a_r| &\ge a_{r, trigger}\\
    |a_r| &\ge a_{r, border}
\end{align}
where $a_{r, trigger} > a_{r, border}$.
For a linear event started at $t_s$ and ended at $t_e$, the valid condition is
\begin{equation}
    \begin{aligned}
        &t_e - t_s \ge t_{linear\_min} \\
        \text{ and } &\min(v_{r, t_s}, v_{r, t_e}) \le v_{stop\_max} \\
        \text{ and } &\max(v_{r, t_s}, v_{r, t_e}) \ge v_{move\_min}
    \end{aligned}
\end{equation}
Then these events are further classified as
\begin{itemize}[\IEEEsetlabelwidth{1000}]
    \item Vehicle starting, if $v_{r, t_s} \le v_{stop\_max}$ \\and $v_{r, t_e} \ge v_{move\_min}$
    \item Vehicle stopping, if $v_{r, t_s} \ge v_{move\_min}$ \\and $v_{r, t_e} \le v_{stop\_max}$
    \item Invalid event, otherwise
\end{itemize}

\subsubsection{Model Parameters}
A total of 11 parameters have been introduced in this model:
\begin{equation}
    \begin{aligned}
        a_{\theta, trigger}, a_{\theta, border}, v_{turn\_min}, t_{turn\_min}, \theta_{min}, \theta_{max} \\
        a_{r, trigger}, a_{r, border}, t_{linear\_min}, v_{stop\_max}, v_{move\_min}
    \end{aligned}
\end{equation}

In a world coordinate system scaled to units of meters, these parameters can be easily set according to common sense.
However, due to detection noise and annotation bias, their value may affect the general performance of event detection.

\subsubsection{Scoring Mechanism}

Unlike training-based systems, our system does not naturally have a scoring mechanism.
However, the evaluation metrics usually require a ranking score for each event.
For the five events, we design a shallow scoring algorithm as 
\begin{itemize}[\IEEEsetlabelwidth{1000}]
    \item Vehicle turning left and vehicle turning right: \\
    \ \ $1 - |\theta - 90| / 90$
    \item Vehicle U-turn: $\theta / 180$
    \item Vehicle starting: $1 - v_{t_s} / v_{stop\_max}$
    \item Vehicle stopping: $1 - v_{t_e} / v_{stop\_max}$
\end{itemize}

\subsection{Traffic collision Detection}

We can also detect traffic collisions. 
Based on distance measurement and overlap of rigid bodies in predicted locations, vehicle crash can be predict. 
Collision scenes are usually more complex and changeable than regular traffic, which makes detection and tracking unreliable.
Such that, we utilize kinematic approaches to detect them beforehand, which also requires no labeled training data.

\subsubsection{Distance Measurement}

With the projected location of vehicles on the ground, we can measure the distances between each pair of vehicles, which are quadrangles.
According to \cite{yu2018traffic}, the minimum distance between two quadrangles $Q_1=A_1B_1C_1D_1$ and $Q2=A_2B_2C_2D_2$ is 
\begin{equation}
    \begin{aligned}
        d_{qq}(Q_1, Q_2) = \min(&\min_{P \in {A_1, B_1, C_1, D_1}}d_{pq}(P, Q_2), \\
        &\min_{P \in {A_2, B_2, C_2, D_2}}d_{pq}(P, Q_1))
    \end{aligned}
\end{equation}
where the minimum distance between a point and a quadrangle $Q=ABCD$ is 
\begin{equation}
    d_{pq}(P, Q) = \min_{e \in {AB, BC, CD, DA}}d_{pe}(P, e)
\end{equation}
A continuous decline of distances between two vehicles indicates a potential crash. 

\subsubsection{Collision Detection}

In this surveillance system, we do not intend to predict crashes ahead of time.
However, we still need to predict the locations of vehicles and therefore detect collision.

Since collision detection only needs prediction for the very near future, we adopt a simple fixed speed kinematic model for location prediction.
With the assumption that a vehicle's shape does not change, we only need to predict its center location $\mathbf{p}$.
Given $\mathbf{p_t} = (x_t, y_t)$, the prediction for $t'$ in the future are calculated by 
\begin{align}
    \hat{x}_{t+t'} &= x_t + v_{x, t} t' \\
    \hat{y}_{t+t'} &= y_t + v_{y, t} t'
\end{align}

Although in the real world, rigid bodies such as vehicles seldom crash into each other to have overlapped locations, they can do so in predictions.
We can check each pair of vehicles for overlaps, which indicates a collision shortly.

\section{Experiments}

\subsection{Implementation Details}

\begin{figure}[!ht]
    \centering
    \includegraphics[width=0.5\linewidth]{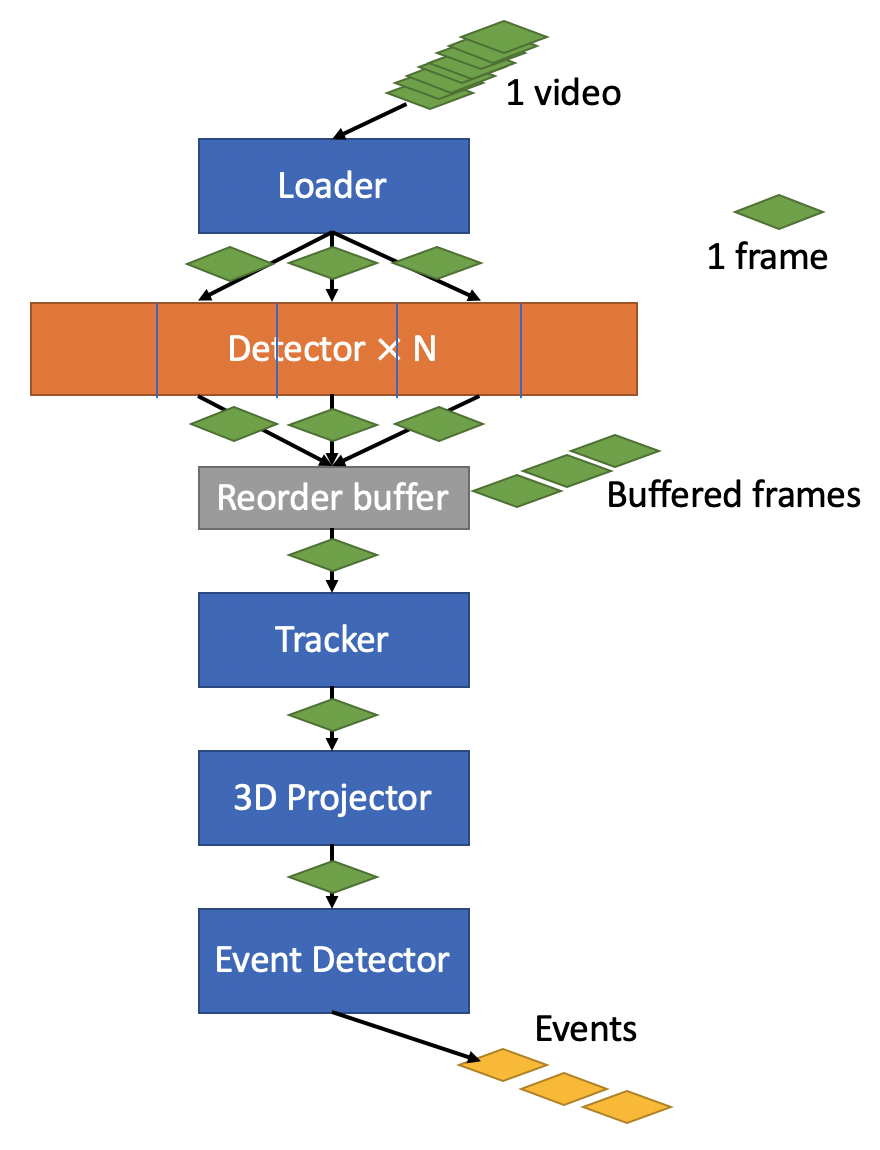}
    \caption{The Real-time System}
    \label{fig/system}
\end{figure}

To deal with the real-world surveillance task, we apply a real-time system which supports stream video execution captured from live surveillance cameras.
As shown in Fig. \ref{fig/system}, the proposed system consists of five submodels, i.e., loader, object detector, tracker, 3D projector, and event detector.

Although providing a reliable performance, the object detector with Mask R-CNN model slows down the system. 
Therefore, a frame-level parallelism schema is applied to overcome the speed bottleneck of the object detector.
The detection results from multiple detectors are reordered in a buffer to go through later stages.

We test our system on several GPU instances independently. 
Each instance contains four Nvidia Geforce RTX 2080Ti GPUs and 128GB memory. 
On average, the system could process the stream video of 1080p resolution at 18fps with a latency of 0.2s.

\subsection{Experimental Study}
\subsubsection{Dataset and Methods}

The Multiview Extended Video with Activities (MEVA)\footnote{\url{http://mevadata.org}} dataset is a currently released large-scale event detection dataset.
Its current release contains 2000 5-minute video clips with 1080p resolution.
However, there is no annotation provided with the raw videos. 
We manually annotate the vehicle actions over the etire dataset. 
The orignal MEVA data are split into a 4870-minute training set and a 500-minute test set. 
In the experiments, we use the full training set and a subset of a 160-minute test set of outdoor scenes with calibrated cameras.

To the best of our knowledge, our system is the first training-free system in the area. 
As a baseline, we re-implement the system introduced in \cite{chen2019minding}. 
It utilizes optical flow features and a Recurrent Neuron Network model for event classification. 
We re-train this system with the training set of MEVA.

The Car Accident Detection and Prediction (CADP) \cite{shah2018cadp} contains traffic collision videos collected from Youtube.
We will use it to test the performance on traffic collision detection.

\subsubsection{Detection and Tracking Result}

In the experiments, all systems utilize the same detection-tracking backbone. 
Therefore, it is necessary to analyze its performance first.
We measure the performance of the detection-tracking model with recall metric. 
It provides an estimation of the upper bound for the down-stream event detection models.

To calculate the recall, we calculate the Intersection of Union (IoU) between 2D bounding boxes of tracked vehicles and event annotations.
Then we match them with the Hungarian algorithm \cite{kuhn1955hungarian}.
Event IoU is calculated as the mean of object IoU over all frames, where a lost detection would result in a 0 IoU.
The recalls of events at different thresholds of event IoU are shown in TABLE \ref{tab/recall}. 
At event IoU threshold 0, the detection-tracking recall reaches the maximum at 0.93.
However, when we lift the threshold to 0.3, it drops to 0.51.
This implies that quite a few lost detection of events are caused by the detection-tracking model, which is not our current focus.
When evaluating and comparing the two systems, we need to take this effect into account.

\begin{table}[!h]
    \caption{Detection and Tracking Result (Recall)}
    \label{tab/recall}
    \centering
    \begin{tabular}{@{}c|cccc@{}}
    \toprule
    Event IoU Threshold         & 0             & 0.1  & 0.2  & 0.3  \\ \midrule
    Vehicle turning left  & 0.95          & 0.78 & 0.59 & 0.62 \\
    Vehicle turning right & 0.91          & 0.71 & 0.59 & 0.35 \\
    Vehicle U-turn        & 0.90          & 0.76 & 0.63 & 0.54 \\
    Vehicle starting      & 0.95          & 0.89 & 0.79 & 0.73 \\
    Vehicle stopping      & 0.90          & 0.79 & 0.74 & 0.62 \\ \midrule
    Average               & 0.93          & 0.78 & 0.65 & 0.51 \\ \bottomrule
    \end{tabular}
\end{table}

\subsubsection{Event Detection Metrics}

The official ActEV Scorer\footnote{\url{https://github.com/usnistgov/ActEV_Scorer}} with protocol ActEV19\_AD\_V2 is used to calculate event detection metrics according to the TRECVID benchmark \cite{awad2018trecvid} and the ActEV SDL evaluation plan\footnote{\url{https://actev.nist.gov/pub/ActEV_SDL_EvaluationPlan_081219.pdf}}.
We use probability of missed detection $P_{miss}$, rate of false alarm $R_{fa}$, and time based false alarm $T_{fa}$ as metrics.
Their definitions are as follows
\begin{align}
    P_{miss} &= \frac{N_{\text{missed\_detections}}}{N_{\text{true\_instances}}} \\
    R_{fa} &= \frac{N_{\text{false\_alarms}}}{\text{video\_duration\_in\_minutes}}
\end{align}
\begin{equation}
    T_{fa} = \frac{1}{\mathit{NR}}\sum_{i=1}^{N_{\text{frames}}} \max (0, D_i - G_i)
\end{equation}
where $\mathit{NR}$ denotes the duration of video without events, $D_i$ the total count of detected events at frame $i$, $G_i$ the total count of groundtruth events at frame $i$.

The general metrics for all events detected by both systems are shown in TABLE \ref{tab/metric_all}.
Concerning $P_{miss}$, our training-free system outperforms the training-based system at almost all events. 
For vehicle starting events, the training-based system gave a lower $P_{miss}$ but a dramatically high $R_{fa}$, which indicates a potential over-fitting.

\begin{table}[h!]
    \caption{General metrics over all detected events\newline(Lower is better for all metrics)}
    \label{tab/metric_all}
    \centering
    \begin{tabular}{@{}c|ccc|ccc@{}}
    \toprule
                          & \multicolumn{3}{c}{Training-free System} & \multicolumn{3}{|c}{Training-based System} \\
                          & $P_{miss}$        & $R_{fa}$      & $T_{fa}$      & $P_{miss}$         & $R_{fa}$       & $T_{fa}$      \\ \midrule
    Vehicle turning left  & {\ul 0.36}     & 0.68     & 0.24     & 0.64            & 1.92      & 0.12     \\
    Vehicle turning right & {\ul 0.47}     & 0.85     & 0.25     & 0.77            & 1.14      & 0.08     \\
    Vehicle U-turn        & {\ul 0.53}     & 0.98     & 0.40     & 1.00            & 0.03      & 0.00     \\
    Vehicle starting      & 0.52           & 0.43     & 0.26     & {\ul 0.32}      & 22.22     & 2.40     \\
    Vehicle stopping      & {\ul 0.68}     & 0.12     & 0.12     & 1.00            & 0.00      & 0.00     \\ \midrule
    Mean                  & \textbf{0.60}  & -        & -        & 0.88            & -         & -        \\ \bottomrule
    \end{tabular}
\end{table}

As a trade-off between Type 1 error and Type 2 error, detected events are sorted according to a score in the original evaluation plan.

\begin{figure}[!ht]
    \centering
    \subfloat{\includegraphics[width=0.48\linewidth]{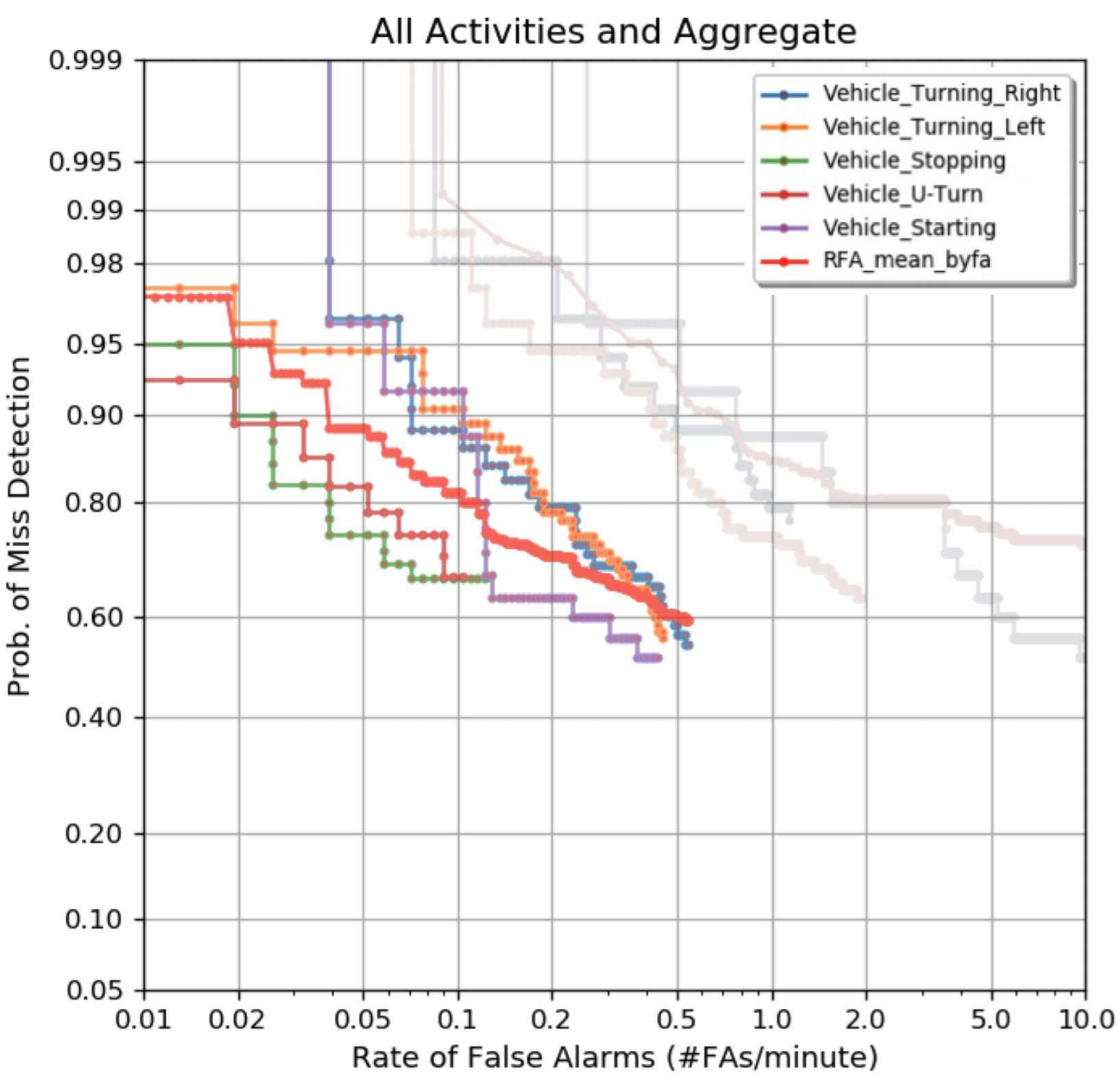}}
    \subfloat{\includegraphics[width=0.48\linewidth]{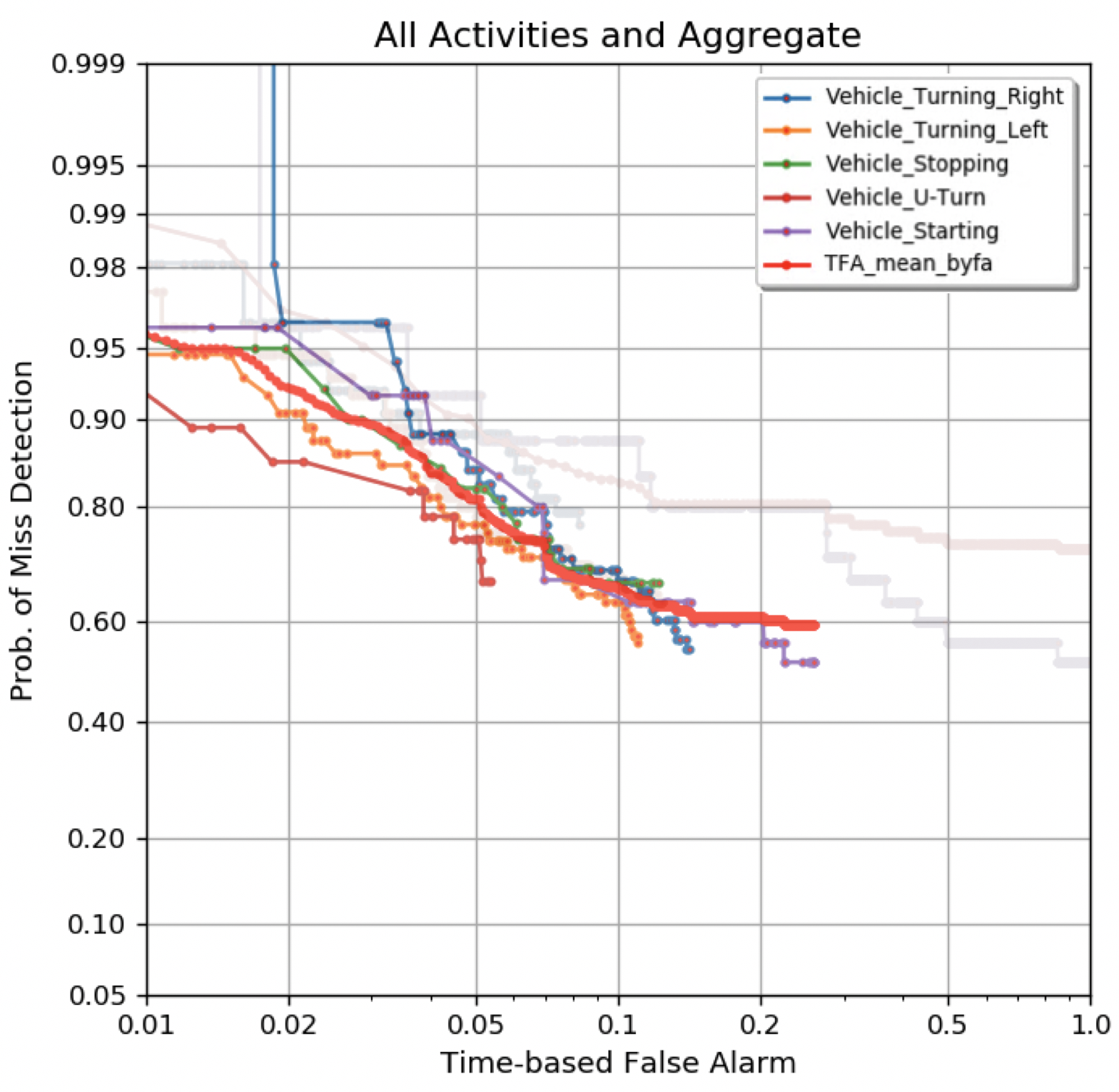}}
    \caption[$P_{miss}$ Curves with $R_{fa}$ (left) and $T_{fa}$ (right)]{$P_{miss}$ curves with $R_{fa}$ (left) and $T_{fa}$ (right). The solid lines are from the training-free model, the transparent lines are from the training-based model.    }
    \label{fig/metric}
\end{figure}

Fig. \ref{fig/metric} shows the curve of p\_miss under different thresholds of $R_{fa}$ and $T_{fa}$.
We can see that our training-free model is significantly better than the training-based model on every level of $R_{fa}$ and $T_{fa}$.
The performance of the training-based system highly relies on the quality and coverage of training data, while our training-free system does not have the problem.
Therefore, our system results in much better performance, while it requires little effort of annotation.

\subsection{Traffic Collision Detection}

\begin{figure}[!h]
    \centering
    \includegraphics[width=0.6\linewidth]{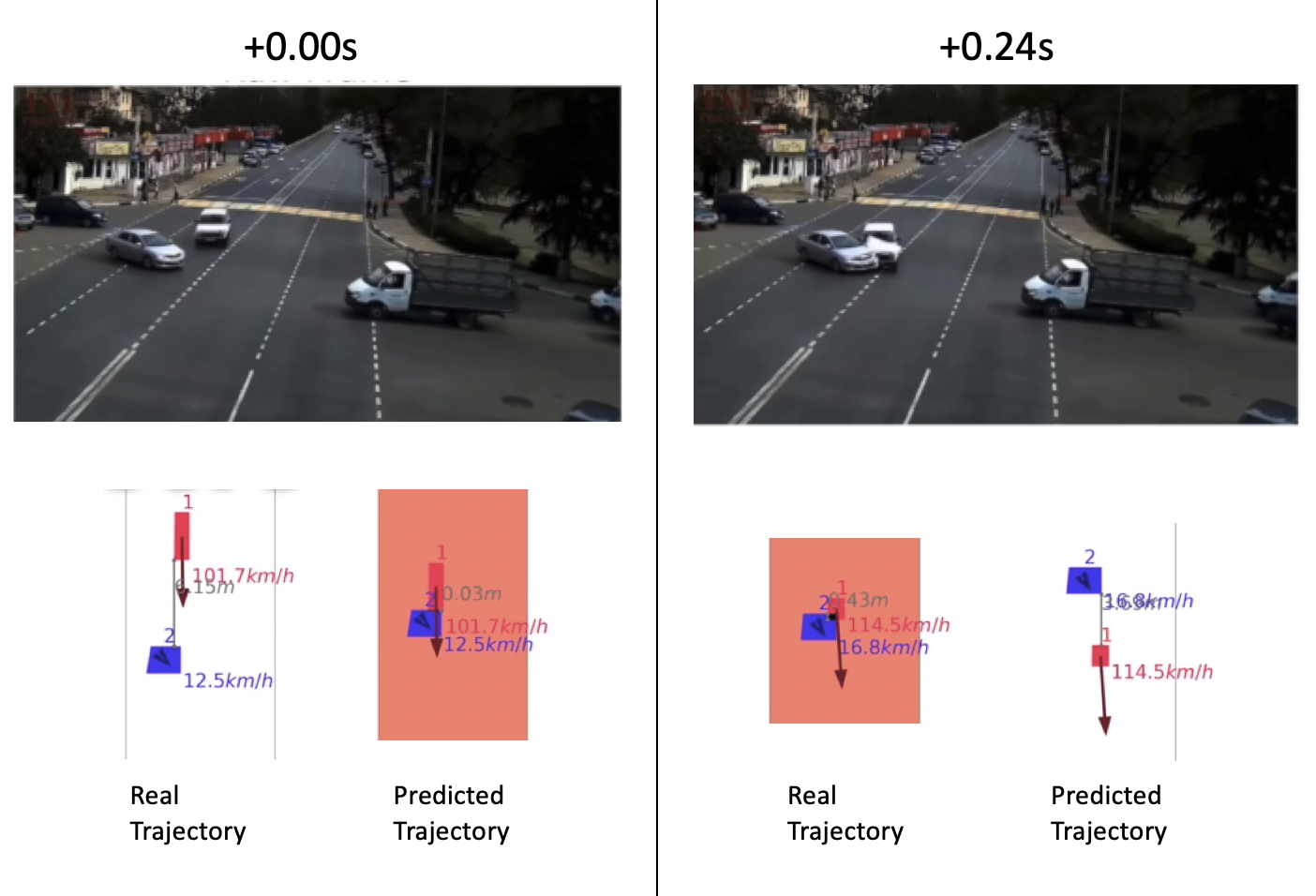}
    \caption[Sample of Crash Detection]{Sample of Crash Detection \\A frame before the crash (left half) and a frame during the crash (right half) are shown. Below the two frames are the real(left) and predicted(right) trajectories in the bird's eye view of the road. Red background in the bird's eye view indicates a detected collision.}
    \label{fig/crash}
\end{figure}

We test our traffic collision detection model on the CADP dataset.
Due to the lack of camera calibration data, we manually annotated parallel lines on a subset of videos to derive the vanishing points.
A sample of crash detection is shown in Fig. \ref{fig/crash}.

The results implies the effectiveness of our system in the collision detection task.
For more samples of the crash detection model, the readers are kindly referred to the link posted in the Abstract section.

\section{Conclusion}

In this paper, we propose a real-time traffic event detection system for traffic surveillance safety.
To the best of our knowledge, this is the first attempt of a training-free system for large-scale traffic event detection.
Our system utilizes a monocular 3D method for surveillance camera views, which overcomes the problem of occlusions and effect of camera viewing angles on conventional 2D methods.
Our system supports real-time stream processing for large-scale traffic data.
It significantly outperforms the existing training-based system on real-world surveillance dataset.

For future works, automatic camera calibration approaches can be developed to extend our system to arbitrary cameras and videos.

\bibliographystyle{IEEEtran}
\bibliography{reference}

\begin{thebibliography}{10}
\providecommand{\url}[1]{#1}
\csname url@samestyle\endcsname
\providecommand{\newblock}{\relax}
\providecommand{\bibinfo}[2]{#2}
\providecommand{\BIBentrySTDinterwordspacing}{\spaceskip=0pt\relax}
\providecommand{\BIBentryALTinterwordstretchfactor}{4}
\providecommand{\BIBentryALTinterwordspacing}{\spaceskip=\fontdimen2\font plus
\BIBentryALTinterwordstretchfactor\fontdimen3\font minus
  \fontdimen4\font\relax}
\providecommand{\BIBforeignlanguage}[2]{{%
\expandafter\ifx\csname l@#1\endcsname\relax
\typeout{** WARNING: IEEEtran.bst: No hyphenation pattern has been}%
\typeout{** loaded for the language `#1'. Using the pattern for}%
\typeout{** the default language instead.}%
\else
\language=\csname l@#1\endcsname
\fi
#2}}
\providecommand{\BIBdecl}{\relax}
\BIBdecl

\bibitem{chen2019minding}
J.~Chen, J.~Liu, J.~Liang, T.-Y. Hu, W.~Ke, W.~Barrios, D.~Huang, and A.~G.
  Hauptmann, ``Minding the gaps in a video action analysis pipeline,'' in
  \emph{2019 IEEE Winter Applications of Computer Vision Workshops
  (WACVW)}.\hskip 1em plus 0.5em minus 0.4em\relax IEEE, 2019, pp. 41--46.

\bibitem{yu2018traffic}
L.~Yu, D.~Zhang, X.~Chen, and A.~Hauptmann, ``Traffic danger recognition with
  surveillance cameras without training data,'' in \emph{2018 15th IEEE
  International Conference on Advanced Video and Signal Based Surveillance
  (AVSS)}.\hskip 1em plus 0.5em minus 0.4em\relax IEEE, 2018, pp. 1--6.

\bibitem{he2017mask}
K.~He, G.~Gkioxari, P.~Doll{\'a}r, and R.~Girshick, ``Mask r-cnn,'' in
  \emph{Proceedings of the IEEE international conference on computer vision},
  2017, pp. 2961--2969.

\bibitem{lin2017feature}
T.-Y. Lin, P.~Doll{\'a}r, R.~Girshick, K.~He, B.~Hariharan, and S.~Belongie,
  ``Feature pyramid networks for object detection,'' in \emph{Proceedings of
  the IEEE conference on computer vision and pattern recognition}, 2017, pp.
  2117--2125.

\bibitem{lin2014microsoft}
T.-Y. Lin, M.~Maire, S.~Belongie, J.~Hays, P.~Perona, D.~Ramanan,
  P.~Doll{\'a}r, and C.~L. Zitnick, ``Microsoft coco: Common objects in
  context,'' in \emph{European conference on computer vision}.\hskip 1em plus
  0.5em minus 0.4em\relax Springer, 2014, pp. 740--755.

\bibitem{wojke2017simple}
N.~Wojke, A.~Bewley, and D.~Paulus, ``Simple online and realtime tracking with
  a deep association metric,'' in \emph{2017 IEEE International Conference on
  Image Processing (ICIP)}.\hskip 1em plus 0.5em minus 0.4em\relax IEEE, 2017,
  pp. 3645--3649.

\bibitem{sochor2017traffic}
J.~Sochor, R.~Jur{\'a}nek, and A.~Herout, ``Traffic surveillance camera
  calibration by 3d model bounding box alignment for accurate vehicle speed
  measurement,'' \emph{Computer Vision and Image Understanding}, vol. 161, pp.
  87--98, 2017.

\bibitem{hartley2003multiple}
R.~Hartley and A.~Zisserman, \emph{Multiple view geometry in computer
  vision}.\hskip 1em plus 0.5em minus 0.4em\relax Cambridge university press,
  2003.

\bibitem{lee2011robust}
S.~C. Lee and R.~Nevatia, ``Robust camera calibration tool for video
  surveillance camera in urban environment,'' in \emph{CVPR 2011
  WORKSHOPS}.\hskip 1em plus 0.5em minus 0.4em\relax IEEE, 2011, pp. 62--67.

\bibitem{dubska2014fully}
M.~Dubsk{\'a}, A.~Herout, R.~Jur{\'a}nek, and J.~Sochor, ``Fully automatic
  roadside camera calibration for traffic surveillance,'' \emph{IEEE
  Transactions on Intelligent Transportation Systems}, vol.~16, no.~3, pp.
  1162--1171, 2014.

\bibitem{suzuki1985topological}
S.~Suzuki \emph{et~al.}, ``Topological structural analysis of digitized binary
  images by border following,'' \emph{Computer vision, graphics, and image
  processing}, vol.~30, no.~1, pp. 32--46, 1985.

\bibitem{sultani2018real}
W.~Sultani, C.~Chen, and M.~Shah, ``Real-world anomaly detection in
  surveillance videos,'' in \emph{Proceedings of the IEEE Conference on
  Computer Vision and Pattern Recognition}, 2018, pp. 6479--6488.

\bibitem{xu2018dual}
Y.~Xu, X.~Ouyang, Y.~Cheng, S.~Yu, L.~Xiong, C.-C. Ng, S.~Pranata, S.~Shen, and
  J.~Xing, ``Dual-mode vehicle motion pattern learning for high performance
  road traffic anomaly detection,'' in \emph{Proceedings of the IEEE Conference
  on Computer Vision and Pattern Recognition Workshops}, 2018, pp. 145--152.

\bibitem{shah2018cadp}
A.~P. Shah, J.-B. Lamare, T.~Nguyen-Anh, and A.~Hauptmann, ``Cadp: A novel
  dataset for cctv traffic camera based accident analysis,'' in \emph{2018 15th
  IEEE International Conference on Advanced Video and Signal Based Surveillance
  (AVSS)}.\hskip 1em plus 0.5em minus 0.4em\relax IEEE, 2018, pp. 1--9.

\bibitem{kuhn1955hungarian}
H.~W. Kuhn, ``The hungarian method for the assignment problem,'' \emph{Naval
  research logistics quarterly}, vol.~2, no. 1-2, pp. 83--97, 1955.

\bibitem{awad2018trecvid}
G.~Awad, A.~Butt, K.~Curtis, Y.~Lee, J.~Fiscus, A.~Godil, D.~Joy, A.~Delgado,
  A.~Smeaton, Y.~Graham \emph{et~al.}, ``Trecvid 2018: Benchmarking video
  activity detection, video captioning and matching, video storytelling linking
  and video search,'' 2018.

\end{thebibliography}

\end{document}